\documentclass[11pt,a4paper]{article}
\usepackage[hyperref]{acl2019}
\usepackage{times}
\usepackage{latexsym}
\usepackage{fancyhdr,graphicx,amsmath,amssymb}
\usepackage[ruled,vlined]{algorithm2e}
\usepackage{url}
\usepackage{array}

\include{pythonlisting}

\aclfinalcopy 

\newcommand\nnfootnote[1]{%
  \begin{NoHyper}
  \renewcommand\thefootnote{}\footnote{#1}%
  \addtocounter{footnote}{-1}%
  \end{NoHyper}
}

\title{Incorporating Word and Subword Units in Unsupervised Machine Translation Using Language Model Rescoring}

\author{Zihan Liu$^*$, Yan Xu$^*$, Genta Indra Winata, Pascale Fung \\
  Center for Artificial Intelligence Research (CAiRE)\\
  Department of Electronic and Computer Engineering\\
  The Hong Kong University of Science and Technology, Clear Water Bay, Hong Kong\\
  \texttt{\{zliucr,yxucb,giwinata\}@connect.ust.hk, pascale@ece.ust.hk}}

\date{}

\begin{document}
\maketitle
\begin{abstract}
This paper describes CAiRE's submission to the unsupervised machine translation track of the WMT'19 news shared task from German to Czech. We leverage a phrase-based statistical machine translation (PBSMT) model and a pre-trained language model to combine word-level neural machine translation (NMT) and subword-level NMT models without using any parallel data. We propose to solve the morphological richness problem of languages by training byte-pair encoding (BPE) embeddings for German and Czech separately, and they are aligned using MUSE \cite{conneau2017word}. To ensure the fluency and consistency of translations, a rescoring mechanism is proposed that reuses the pre-trained language model to select the translation candidates generated through beam search. Moreover, a series of pre-processing and post-processing approaches are applied to improve the quality of final translations.
\end{abstract}

\section{Introduction}
\nnfootnote{*These two authors contributed equally.}
Machine translation (MT) has achieved huge advances in the past few years~\cite{bahdanau2014neural,gehring2017convolutional,vaswani2017attention,vaswani2018tensor2tensor}. However, the need for a large amount of manual parallel data obstructs its performance under low-resource conditions. Building an effective model on low resource data or even in an unsupervised way is always an interesting and challenging research topic \cite{gu2018universal, radford2015unsupervised, lee2019team}. Recently, unsupervised MT~\cite{artetxe2017unsupervised,artetxe2018unsupervised,conneau2017word,lample2018phrase,wu2019extract}, which can immensely reduce the reliance on parallel corpora, has been gaining more and more interest. 

Training cross-lingual word embeddings~\cite{conneau2017word,artetxe2017learning} is always the first step of the unsupervised MT models which produce a word-level shared embedding space for both the source and target, but the lexical coverage can be an intractable problem. To tackle this issue, ~\citet{sennrich2016neural} provided a subword-level solution to overcome the out-of-vocabulary (OOV) problem. 

In this work, the systems we implement for the German-Czech language pair are built based on the previously proposed unsupervised MT systems, 
with some adaptations made to accommodate the morphologically rich characteristics of German and Czech~\cite{tsarfaty2010statistical}. 
Both word-level and subword-level neural machine translation (NMT) models are applied in this task and further tuned by pseudo-parallel data generated from a phrase-based statistical machine translation (PBSMT) model, which is trained following the steps proposed in~\citet{lample2018phrase} without using any parallel data.
We propose to train BPE embeddings for German and Czech separately and align those trained embeddings into a shared space with MUSE \cite{conneau2017word} to reduce the combinatorial explosion of word forms for both languages. To ensure the fluency and consistency of translations, an additional Czech language model is trained to select the translation candidates generated through beam search by rescoring them. Besides the above, a series of post-processing steps are applied to improve the quality of final translations. Our contribution is two-fold:
\begin{itemize}
    \item We propose a method to combine word and subword (BPE) pre-trained input representations aligned using MUSE \cite{conneau2017word} as an NMT training initialization on a morphologically-rich language pair such as German and Czech.
    \item We study the effectiveness of language model rescoring to choose the best sentences and unknown word replacement (UWR) procedure to reduce the drawback of OOV words.
\end{itemize}

This paper is organized as follows: in Section \ref{section:method}, we describe our approach to the unsupervised translation from German to Czech. Section \ref{section:result} reports the training details and the results for each steps of our approach. More related work is provided in Section \ref{section:related}. Finally, we conclude our work in Section \ref{section:concl}. 

\section{Methodology} 
\label{section:method}

\begin{figure*}[ht!]
\centering
\includegraphics[scale=0.15]{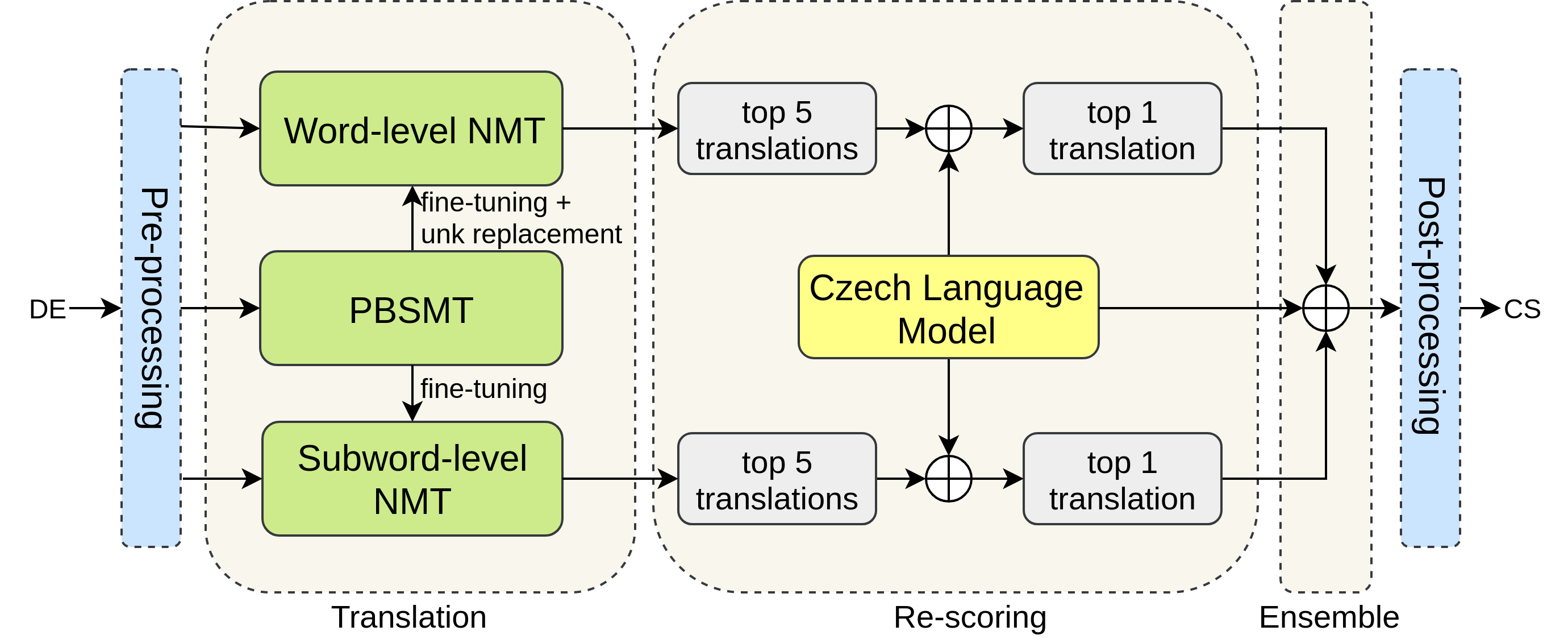}
\caption{The illustration of our system. The translation procedure can be divided into five steps: (a) pre-processing, (b) translation generation (\S \ref{sec:umt}) from word-level NMT, subword-level NMT, and PBSMT. In the training, we fine-tune word-level and subword-level NMT models with pseudo-parallel data from NMT models and the best PBSMT model. Moreover, an unknown word replacement mechanism (\S \ref{sec:unk}) is applied to the translations generated from the word-level NMT model, (c) translation candidate rescoring, (d) construction of an ensemble of the translations from NMT models, and (e) post-processing. }
\label{fig:system}
\end{figure*}

In this section, we describe how we built our main unsupervised machine translation system, which is illustrated in Figure \ref{fig:system}. 

\subsection{Unsupervised Machine Translation} \label{sec:umt}
\subsubsection{Word-level Unsupervised NMT} \label{sec:wlunmt}
We follow the unsupervised NMT in \citet{lample2018phrase} by leveraging initialization, language modeling and back-translation. However, instead of using BPE, we use MUSE \cite{conneau2017word} to align word-level embeddings of German and Czech, which are trained by FastText \cite{bojanowski2017enriching} separately. We leverage the aligned word embeddings to initialize our unsupervised NMT model. 

The language model is a denoising auto-encoder, which is trained by reconstructing original sentences from noisy sentences. The process of language modeling can be expressed as minimizing the following loss:
\begin{align*}
    L^{lm} = \lambda * \{ & E_{x\sim S}[-log P_{s \to s}(x|N(x))] + \\
                          & E_{y\sim T}[-log P_{t \to t}(x|N(y))] \},  \tag{1} \label{eq1}
\end{align*}
where $N$ is a noise model to drop and swap some words with a certain probability in the sentence $x$, $P_{s \to s}$ and $P_{t \to t}$ operate on the source and target sides separately, and $\lambda$ acts as a weight to control the loss function of the language model.
a
Back-translation turns the unsupervised problem into a supervised learning task by leveraging the generated pseudo-parallel data. The process of back-translation can be expressed as minimizing the following loss:
\begin{align*}
    L^{bt} = & E_{x\sim S}[-log P_{t \to s}(x|v^*(x))] + \\
             & E_{y\sim T}[-log P_{s \to t}(y|u^*(y))], \tag{2}
\end{align*}
where $v^*(x)$ denotes sentences in the target language translated from source language sentences $S$, $u^*(y)$ similarly denotes sentences in the source language translated from the target language sentences $T$ and $P_{t \to s}$, and $P_{s \to t}$ denote the translation direction from target to source and from source to target respectively. 

\subsubsection{Subword-level Unsupervised NMT}
We note that both German and Czech~\cite{tsarfaty2010statistical} are morphologically rich languages, which leads to a very large vocabulary size for both languages, but especially for Czech (more than one million unique words for German, but three million unique words for Czech). 
To overcome OOV issues, we leverage subword information, which can lead to better performance.

We employ subword units \cite{sennrich2016improving} to tackle the morphological richness problem. There are two advantages of using the subword-level. First, we can alleviate the OOV issue by zeroing out the number of unknown words. Second, we can leverage the semantics of subword units from these languages. However, German and Czech are distant languages that originate from different roots, so they only share a small fraction of subword units. To tackle this problem, we train FastText word vectors \cite{bojanowski2017enriching} separately for German and Czech, and apply MUSE \cite{conneau2017word} to align these embeddings. 

\subsubsection{Unsupervised PBSMT}
PBSMT models can outperform neural models in low-resource conditions. A PBSMT model utilizes a pre-trained language model and a phrase table with phrase-to-phrase translations from the source language to target languages, which provide a good initialization. 
The phrase table stores the probabilities of the possible target phrase translations corresponding to the source phrases, which can be referred to as $P(s|t)$, with $s$ and $t$ representing the source and target phrases. The source and target phrases are mapped according to inferred cross-lingual word embeddings, which are trained with monolingual corpora and aligned into a shared space without any parallel data~\cite{artetxe2017learning,conneau2017word}. 

We use a pre-trained n-gram language model to score the phrase translation candidates by providing the relative likelihood estimation $P(t)$, so that the translation of a source phrase is derived from: $arg max_{t} P(t|s)=arg max_{t} P(s|t)P(t)$.

Back-translation enables the PBSMT models to be trained in a supervised way by providing pseudo-parallel data from the translation in the reverse direction, which indicates that the PBSMT models need to be trained in dual directions so that the two models trained in the opposite directions can promote each other's performance.

In this task, we follow the method proposed by \citet{lample2018phrase} to initialize the phrase table, train the KenLM language models~\cite{heafield2011kenlm}\footnote{The code can be found at https://github.com/kpu/kenlm} and train a PBSMT model, but we make two changes. First, we only initialize a uni-gram phrase table because of the large vocabulary size of German and Czech and the limitation of computational resources. Second, instead of training the model in the \textit{truecase} mode, we maintain the same pre-processing step (see more details in \S \ref{preproc}) as the NMT models.

\subsubsection{Fine-tuning NMT}
We further fine-tune the NMT models mentioned above on the pseudo-parallel data generated by a PBSMT model. We choose the best PBSMT model and mix the pseudo-parallel data from the NMT models and the PBSMT model, which are used for back-translation. The intuition is that we can use the pseudo-parallel data produced by the PBSMT model as the supplementary translations in our NMT model, and these can potentially boost the robustness of the NMT model by increasing the variety of back-translation data. 

\subsection{Unknown Word Replacement} \label{sec:unk}

\begin{figure*}[ht!]
\centering
\includegraphics[scale=0.21]{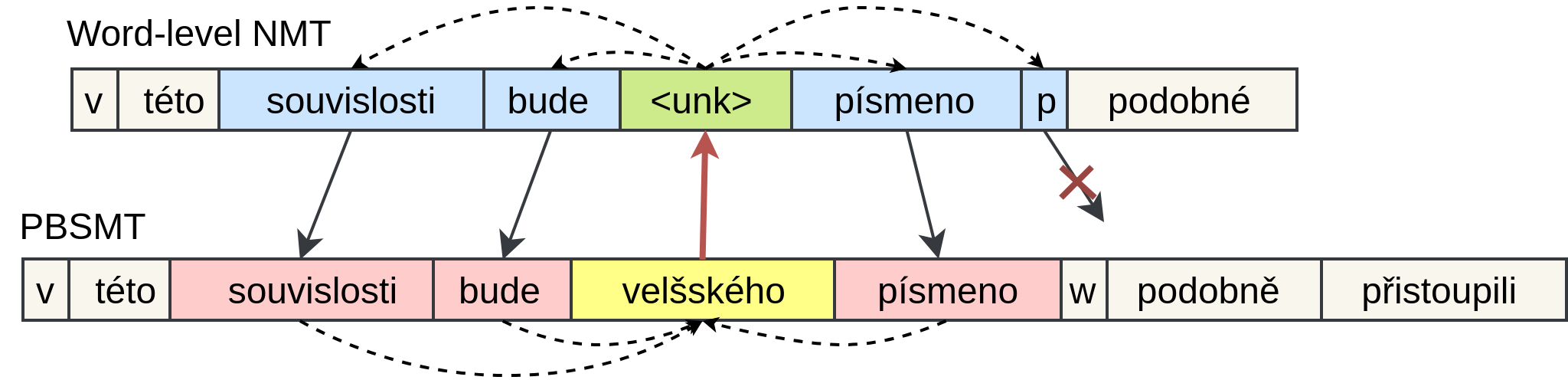}
\caption{The illustration of the unknown word replacement (UWR) procedure for word-level NMT. The words of the PBSMT model translation in the pink boxes match the context words of the unknown word \texttt{<UNK>} in the word-level NMT model translation in the blue boxes. Finally, we choose a possible target word, in the yellow box, from the PBSMT model translation to replace the unknown word in the green box.}
\label{fig:unk}
\end{figure*}

Around 10\% of words found in our NMT training data are unknown words (\texttt{<UNK>}), which immensely limits the potential of the word-level NMT model.
In this case, replacing unknown words with reasonable words can be a good remedy. Then, assuming the translations from the word-level NMT model and PBSMT model are roughly aligned in order, we can replace the unknown words in the NMT translations with the corresponding words in the PBSMT translations. Compared to the word-level NMT model, the PBSMT model ensures that every phrase will be translated without omitting any pieces from the sentences. We search for the word replacement by the following steps, which are also illustrated in Figure \ref{fig:unk}:
\paragraph{\textbf{Step 1}}
For every unknown word, we can get the context words with a context window size of two.
\paragraph{\textbf{Step 2}}
Each context word is searched for in the corresponding PBSMT translation. From our observation, the meanings of the words in Czech are highly likely to be the same if only the last few characters are different. Therefore, we allow the last two characters to be different between the context words and the words they match.
\paragraph{\textbf{Step 3}}
If several words in the PBSMT translation match a context word, the word that is closest to the position of the context word in the PBSMT translation will be selected and put into the candidate list to replace the corresponding \texttt{<UNK>} in the translation from the word-level NMT model.
\paragraph{\textbf{Step 4}}
Step 2 and Step 3 are repeated until all the context words have been searched. After removing all the punctuation and the context words in the candidate list, the replacement word is the one that most frequently appears in the candidate list. If no candidate word is found, we just remove the \texttt{<UNK>} without adding a word.


\subsection{Language Model Rescoring} \label{sec:lm}

Instead of direct translation with NMT models, we generate several translation candidates using beam search with a beam size of five. We build the language model proposed by \citet{merityRegOpt,merityAnalysis} trained using a monolingual Czech dataset to rescore the generated translations. The scores are determined by the perplexity (PPL) of the generated sentences and the translation candidate with the lowest PPL will be selected as the final translation.

\subsection{Model Ensemble} \label{sec:ensemble}
Ensemble methods have been shown very effective in many natural language processing tasks. We apply an ensemble method by taking the top five translations from word-level and subword-level NMT, and rescore all translations using
our pre-trained Czech language model mentioned in \S \ref{sec:lm}. Then, we select the best translation with the lowest perplexity.

\section{Experiments} \label{section:result}

\subsection{Data Pre-processing} \label{preproc}
We note that in the corpus, there are tokens representing quantity or date. Therefore, we delexicalize the tokens using two special tokens: (1) \texttt{<NUMBER>} to replace all the numbers that express a specific quantity, and (2) \texttt{<DATE>} to replace all the numbers that express a date. Then, we retrieve these numbers in the post-processing. 
There are two advantages of data pre-processing. First, replacing numbers with special tokens can reduce vocabulary size. Second, the special tokens are more easily processed by the model. 

\subsection{Data Post-processing} \label{sec:postproc}
\paragraph{Special Token Replacement}
In the pre-processing, we use the special tokens \texttt{<NUMBER>} and \texttt{<DATE>} to replace numbers that express a specific quantity and date respectively. Therefore, in the post-processing, we need to restore those numbers. We simply detect the pattern \texttt{<NUMBER>} and \texttt{<DATE>} in the original source sentences and then replace the special tokens in the translated sentences with the corresponding numbers detected in the source sentences. In order to make the replacement more accurate, we will detect more complicated patterns like \texttt{<NUMBER> / <NUMBER>} in the original source sentences. If the translated sentences also have the pattern, we replace this pattern \texttt{<NUMBER> / <NUMBER>} with the corresponding numbers in the original source sentences.

\paragraph{Quotes Fixing}
The quotes are fixed to keep them the same as the source sentences.

\paragraph{Recaser}
For all the models mentioned above that work under a lower-case setting, a recaser implemented with Moses~\cite{koehn2007moses} is applied to convert the translations to the real cases. 

\paragraph{Patch-up}
From our observation, the ensemble NMT model lacks the ability to translate name entities correctly. 
We find that words with capital characters are named entities, and those named entities in the source language may have the same form in the target language. Hence, we capture and copy these entities at the end of the translation if they does not exist in our translation.

\subsection{Training}
\paragraph{Unsupervised NMT}
The settings of the word-level NMT and subword-level NMT are the same, except the vocabulary size. We use a vocabulary size of 50k in the word-level NMT setting and 40k in the subword-level NMT setting for both German and Czech. In the encoder and decoder, we use a transformer \cite{vaswani2017attention} with four layers and a hidden size of 512. We share all encoder parameters and only share the first decoder layer across two languages to ensure that the latent representation of the source sentence is robust to the source language.
We train auto-encoding and back-translation during each iteration. As the training goes on, the importance of language modeling become a less important compared to back-translation. Therefore the weight of auto-encoding ($\lambda$ in equation (\ref{eq1})) is decreasing during training.

\paragraph{Unsupervised PBSMT}
The PBSMT is implemented with Moses using the same settings as those in \citet{lample2018phrase}. The PBSMT model is trained iteratively. Both monolingual datasets for the source and target languages consist of 12 million sentences, which are taken from the latest parts of the WMT monolingual dataset. At each iteration, two out of 12 million sentences are randomly selected from the the monolingual dataset. 

\paragraph{Language Model}
According to the findings in \citet{cotterell2018all}, the morphological richness of a language is closely related to the performance of the model, which indicates that the language models will be extremely hard to train for Czech, as it is one of the most complex languages.
We train the QRNN model with 12 million sentences randomly sampled from the original WMT Czech monolingual dataset,~\footnote{http://www.statmt.org/wmt19/} which is also pre-processed in the way mentioned in \S \ref{preproc}. To maintain the quality of the language model, we enlarge the vocabulary size to three million by including all the words that appear more than 15 times. Finally, the PPL of the language model on the test set achieves 93.54. 

\paragraph{Recaser}
We use the recaser model provided in Moses and train the model with the two million latest sentences in the Czech monolingual dataset. After the training procedure, the recaser can restore words to the form in which the maximum probability occurs.

\subsection{PBSMT Model Selection}
The BLEU (cased) score of the initialized phrase table and models after training at different iterations are shown in Table \ref{tab:smt}. From comparing the results, we observe that back-translation can improve the quality of the phrase table significantly, but after five iterations, the phrase table has hardly improved. The PBSMT model at the sixth iteration is selected as the final PBSMT model.
\begin{table}[!htbp]
\begin{center}
\begin{tabular}{lc}
\hline
Model & BLEU Cased\\
\hline\hline
Unsupervised PBSMT &\\
\hline
Unsupervised Phrase Table   & 3.8\\
\quad+ Back-translation Iter. 1   & 6.6\\
\quad+ Back-translation Iter. 2   & 7.3\\
\quad+ Back-translation Iter. 3   & 7.5\\
\quad+ Back-translation Iter. 4   & 7.6\\
\quad+ Back-translation Iter. 5   & 7.7\\
\quad+ Back-translation Iter. 6   & \textbf{7.7}\\
\hline
\end{tabular}
\end{center}
\caption{Results of PBSMT at different iterations}
\label{tab:smt}
\end{table}

\begin{table*}[!htbp]
\begin{center}
\begin{tabular}{lccccc}
\hline
\textbf{Model} & \textbf{BLEU} & \textbf{BLEU Cased} & \textbf{TER} & \textbf{BEER 2.0} & \textbf{CharacterTER}\\
\hline\hline
Unsupervised PBSMT &&&&&\\
\hline
Unsupervised phrase table   & 4   & 3.8 & -     & 0.384 & 0.773\\
\quad+ Back-translation Iter. 6   & 8.3 & 7.7 & 0.887 & 0.429 & \textbf{0.743} \\
\hline\hline
\multicolumn{6}{l}{Unsupervised NMT}\\
\hline
Subword-level NMT      & 9.4 & 9.1 & -   & 0.419 & 0.756\\
\quad+ fine-tuning     & 9.8 & 9.5 & 0.832 & 0.424 & 0.756\\
\quad+ fine-tuning + rescoring & 10.3 & 10 & 0.833 & 0.426 & 0.749\\
\hline
Word-level NMT         & 7.9 & 7.6 & -   & 0.412 & 0.823\\
\quad+ fine-tuning     & 7.9 & 7.7 & -   & 0.413 & 0.819\\
\quad+ fine-tuning + UWR & 10.1 & 9.6 & \textbf{0.829} & \textbf{0.432} & 0.766\\
\quad+ fine-tuning + UWR + rescoring & 10.4 & 9.9 & \textbf{0.829} & \textbf{0.432} & 0.764\\
\hline\hline
\multicolumn{6}{l}{Model Ensemble}\\
\hline
Best Word-level + Subword-level & \textbf{10.6} & \textbf{10.2} & \textbf{0.829} & 0.429 & 0.755\\
\quad+ patch-up            & \textbf{10.6} & \textbf{10.2} & 0.833 & 0.430 & 0.757\\
\hline
\end{tabular}
\end{center}
\caption{Unsupervised translation results. We report the scores of several evaluation methods for every step of our approach. Except the result that is listed on the last line, all results are under the condition that the translations are post-processed without patch-up.}
\label{tab:bleu}
\end{table*}

\subsection{Results}

The performances of our final model and other baseline models are illustrated in Table \ref{tab:bleu}. In the baseline unsupervised NMT models, subword-level NMT outperforms word-level NMT by around a 1.5 BLEU score. Although the unsupervised PBSMT model is worse than the subword-level NMT model, leveraging generated pseudo-parallel data from the PBSMT model to fine-tune the subword-level NMT model can still boost its performance. 
However, this pseudo-parallel data from the PBSMT model can not improve the word-level NMT model since the large percentage of OOV words limits its performance. After applying unknown words replacement to the word-level NMT model, the performance improves by a BLEU score of around 2. Using the Czech language model to re-score helps the model improve by around a 0.3 BLEU score each time. We also use this language model to create an ensemble of the best word-level and subword-level NMT model and achieve the best performance.

\section{Related Work} \label{section:related}
\subsection{Unsupervised Cross-lingual Embeddings}
Cross-lingual word embeddings can provide a good initialization for both the NMT and SMT models. In the unsupervised senario, \citet{artetxe2017learning} independently trained embeddings in different languages using monolingual corpora, and then learned a linear mapping to align them in a shared space based on a bilingual dictionary of a negligibly small size. \citet{conneau2017word} proposed a fully unsupervised learning method to build a bilingual dictionary without using any foregone word pairs, but by considering words from two languages that are near each other as pseudo word pairs.
\citet{lample2019cross} showed that cross-lingual language model pre-training can learn a better cross-lingual embeddings to initialize an unsupervised machine translation model.

\subsection{Unsupervised Machine Translation}
In \citet{artetxe2017unsupervised} and \citet{lample2018unsupervised}, the authors proposed the first unsupervised machine translation models which combines an auto-encoding language model and back-translation in the training procedure. \citet{lample2018phrase} illustrated that initialization, language modeling, and back-translation are key for both unsupervised neural and statistical machine translation. \citet{artetxe2018unsupervised} combined back-translation and MERT \cite{och2003minimum} to iteratively refine the SMT model. \citet{wu2019extract} proposed to discard back-translation. Instead, they extracted and edited the nearest sentences in the target language to construct pseudo-parallel data, which was used as a supervision signal.

\section{Conclusion} \label{section:concl}
In this paper, we propose to combine word-level and subword-level input representation in unsupervised NMT training on a morphologically rich language pair, German-Czech, without using any parallel data. Our results show the effectiveness of using language model rescoring to choose more fluent translation candidates. A series of pre-processing and post-processing approaches improve the quality of final translations, particularly to replace unknown words with possible relevant target words.

\section*{Acknowledgments}
We would like to thank our colleagues Jamin Shin, Andrea Madotto, and Peng Xu for insightful discussions. This work has been partially funded by ITF/319/16FP and MRP/055/18 of the Innovation Technology Commission, the Hong Kong SAR Government.

\newpage

\bibliography{acl2019}
\bibliographystyle{acl_natbib}

\end{document}